\documentclass[letterpaper, 10 pt, journal, twoside]{ieeetran} 
\usepackage{graphics}
\usepackage{epsfig}
\usepackage{mathptmx}
\usepackage{times}
\usepackage{amsmath}
\usepackage{amssymb}
\usepackage{relsize}
\usepackage{subfigure} 
\usepackage{algorithm}
\usepackage{algorithmic}
\usepackage{url}
\DeclareMathAlphabet{\pazocal}{OMS}{zplm}{m}{n}
\usepackage[noadjust]{cite}

\title{Dynamical System Segmentation for Information Measures in Motion}
\author{Thomas A. Berrueta, Ana Pervan, Kathleen Fitzsimons, and Todd D. Murphey%
	\thanks{Manuscript received: July 18, 2018; Revised: October 24, 2018; Accepted: November 15, 2018.
	}%
	\thanks{This work was supported by the NSF under grant CBET-1637764 and by the NDSEG Graduate Fellowship program.
	}%
	\thanks{Authors are with the Neuroscience and Robotics Laboratory (NxR) at the Department of Mechanical Engineering, Northwestern University,
		Evanston, IL. Email:
		{\tt\footnotesize \{tberrueta, anapervan, k-fitzsimons\} @u.northwestern.edu},
		{\tt\footnotesize t-murphey@northwestern.edu}.
	}%
	\thanks{Digital Object Identifier (DOI): see top of this page.}%
}

\begin{document}
	\maketitle
	\begin{abstract}
	Motions carry information about the underlying task being executed. Previous work in human motion analysis suggests that complex motions may result from the composition of fundamental submovements called movemes. The existence of finite structure in motion motivates information-theoretic approaches to motion analysis and robotic assistance. We define task embodiment as the amount of task information encoded in an agent's motions. By decoding task-specific information embedded in motion, we can use task embodiment to create detailed performance assessments. We extract an alphabet of behaviors comprising a motion without \textit{a priori} knowledge using a novel algorithm, which we call dynamical system segmentation. For a given task, we specify an optimal agent, and compute an alphabet of behaviors representative of the task. We identify these behaviors in data from agent executions, and compare their relative frequencies against that of the optimal agent using the Kullback-Leibler divergence. We validate this approach using a dataset of human subjects ($\mathbf{n=53}$) performing a dynamic task, and under this measure find that individuals receiving assistance better embody the task. Moreover, we find that task embodiment is a better predictor of assistance than integrated mean-squared-error.
	\end{abstract}
	
	\section{Introduction}
	\IEEEPARstart{M}{otion} signals encode information about the underlying task being executed, yet the form this information takes may vary. Typically, we represent motion using continuous real-valued signals. While this representation can provide detailed descriptions of an agent's motions, it can be cumbersome. However, based on our choice of representation, we can compress motion signals while preserving information about the task \cite{Shannon}.                           
	
	In \cite{MussaIvaldi1755}, the authors propose that human motions are the result of the composition of a finite set of premotor signals emanating from the spinal cord. As a consequence, the neurological feasibility of motion decomposition forms the basis for action in movement primitives, also known as \emph{movemes} \cite{DELVECCHIO200225}. Movemes are fundamental units of motion, and derive their name from their linguistic analogue: phonemes. Thus, all smooth human motions may be comprised of symbolic sequences drawn from an alphabet of movemes. Movemes motivate the application of information measures in human motion analysis, because they provide evidence of finite structure in otherwise continuous motion signals. Moreover, the existence of movemes indicates that under some choice of representation human motion can be discretized without loss of information. 
	
	In the human motion analysis literature, movemes are often characterized using causal dynamical systems \cite{DELVECCHIO200225,DELVECCHIO20032085}, or hybrid system identification methods, such as autoregressive models \cite{GONZALEZ2015291}. Most motor signal segmentation methods demand \textit{prior specification} of the moveme alphabet either through direct template matching or manual labeling of training data, which limits their use in exploratory analyses where the structure of the alphabet may not be known \textit{a priori}. Techniques in symbolic dynamic filtering can generate symbolic alphabets by creating partitions of the state-space using methods such as maximum entropy partitioning \cite{Ray2008}. Additionally, state-space partition techniques can be applied to nonlinear transformations of the space via methods such as wavelet transforms \cite{Rao2009}. However, the symbols synthesized by these spatial techniques are quasi-static, and are not designed to describe the dynamic nature of movemes. 
	
	Automatic segmentation methods based on Hidden Markov Models (HMMs) describe complex motion as the stochastic evolution of discrete hidden states. Pure HMM-based approaches have been used to directly classify movemes from sequential observations. States in the HMM learn an observation distribution that describes each moveme. Movement primitive HMMs (MP-HMMs) model the temporal phases of an individual primitive and have been used to assess movement performance and generate detailed models of motion \cite{HMM}. Since the model's hidden states are represented by observation distributions, typical HMMs and MP-HMMs---as well as their higher-order and hierarchical variants---all use static symbols, which limits their use in characterizing movemes.
	
	Alternatively, Switched Linear Dynamical Systems (SLDS) partition nonlinear systems into a set of piecewise-continuous linear dynamical systems, whose transitions are modeled by the stochastic dynamics of a hidden switching variable described by an HMM \cite{Pavlovic2000}. Movemes in an SLDS model are represented by linear dynamical systems, which ensures the symbolic alphabet is comprised of dynamic elements. In \cite{Baptista2017}, the authors apply SLDS to model discrete movements, and generate automatic segmentations after training on a manually labeled dataset. While these methods give a trained user tools to analyze motor performance, they rely on the user's ability to identify movemes by themselves or with the aid of a template. This is particularly an issue for analyses of atypical or impaired motion where asymmetries arise, and signals may not match a given template with standard features. Thus, supervised learning approaches are generally not as effective when the structure of the symbolic alphabet is not known ahead of time.
	
	In order to address cases with an unknown number of symbols in the alphabet, the authors of \cite{Vidal2005} introduced a recursive identification framework using switched autoregressive systems without presupposing the number of models required, while work by \cite{Willsky2010} established a similar algorithm based on Bayesian inference over SLDS model parameters given some prior belief. Additionally, unsupervised learning techniques have been employed for decades in the fields of image processing and computer vision to find and label key features in dense sources of data such as video \cite{Zhuang1998,Clarkson1999,Gunsel1998}. Within human motion analysis, motion capture setups are a common means of data acquisition. In this setting, unsupervised segmentations of human motion capture data based on cascading linear dynamical systems may correspond to the identification of movemes from video sequences \cite{Turaga2009}. However, by modeling the switches between movemes with exclusively temporal dependencies (i.e. with Markovian dynamics), any state-space dependence on the switching conditions between movemes is unmodeled, failing to provide transition guard conditions.
	
	\textit{Our approach synthesizes finite sets of dynamic symbols from agents' motions without any prior system knowledge, while modeling state-space dependencies between symbols.} Motivated by movemes and movement primitives, we define \emph{behaviors} as moveme analogues for general systems, and specify them using finite-dimensional nonlinear causal dynamical systems. Making a choice of representation for behaviors is very important. While there exist many data-driven function approximation methods, we choose the Koopman operator to represent behaviors because they are capable of capturing nonlinearities within a linear systems framework \cite{original_koopman}. By identifying features from motion signals and constructing an alphabet of behaviors, we directly encode task-specific information into the symbolic representation. 
	
	Information is an ordered sequence of symbols drawn from an alphabet generated by a source \cite{Shannon}. We define task information with respect to a source, such that the source generates sequences of symbols comprising realizations of the task. Any physical system or agent attempting a given task is a task-specific information source. Throughout this study, we represent task information using the distribution of relative frequencies of symbols within a given set of realizations of the underlying system. Additionally, we define \textit{task embodiment} as a measure of task information encoded in an agent's trajectories by calculating their relative entropy with respect to a reference symbol distribution.
	
	The task embodiment formalism is important because it is agnostic to specification of the system, task, or symbolic representation, which allows us to analyze motion signals generally. Task embodiment provides a framework for making motor performance assessments in tasks where measures such as error are unsuitable. Additionally, the proposed unsupervised segmentation technique is capable of generating symbolic partitions of complex, continuous movements that are not amenable to supervised or template-based approaches---even when the movements are atypical and asymmetrical. Consider human walking as an example. Gait phase partitioning is a canonical problem in the motor segmentation literature \cite{Taborri2010,Ting2018}. Supervised segmentation methods work well for analyzing healthy gaits due to an abundance of data and clinically verified motion templates. Since most impairments are unique there are no equivalent databases for atypical gaits, demanding unsupervised techniques to model gaits such as those proposed in this work. Detailed models of atypical motion can facilitate the development of sophisticated robotic assistance, including methods for exoskeleton-assisted gait. Additionally, performance assessments of gaits typically use heuristic-based approaches. Typical notions of error are ill-suited to comparing gaits since they depend on analyzing joint trajectories which do not easily generalize from one individual to another. However, gait cycles could be systematically compared by using measures of task embodiment.
	
	The primary contributions of this paper are the following. First, we develop a methodology for data-driven partitioning of dynamical systems. These partitions are projections onto the state-space that can be used to extract an alphabet of system behaviors, and can be represented by a graph. Second, we demonstrate that by tracking relative frequencies of behaviors we can discern relationships in human motion, such as whether an individual is receiving task assistance. We apply our information-theoretic approach to a dataset of human subjects $(n=53)$ performing a dynamic task where assistance is sometimes provided, and extract an alphabet of optimal behaviors based on a synthesized exemplar agent. By tracking the relative frequencies of finite behaviors in human subjects and comparing to those of the optimal agent, we are able to quantify the degree of task embodiment, and determine whether a subject received assistance. We validate the performance of task embodiment by using integrated mean-squared-error (MSE) as a baseline, and find that task embodiment is a good predictor of assistance.
	
	\section{Methods}
	An agent's state trajectories simultaneously encode information about the system dynamics and the task it executes. By examining system trajectories, one can uncover patterns in how it traverses the underlying state-space manifold. We propose Dynamical System Segmentation (DSS): a nonparametric, unsupervised, data-driven algorithm for creating low-dimensional, graphical representations of system behaviors by generating partitions of the state-space manifold sensitive to the underlying distribution of task information.

	\subsection{Koopman Operators}
	We use Koopman operators as our choice of representation for system behaviors because they are capable of representing nonlinear systems within a compact, linear form amenable to control hierarchies. Consider dynamical systems described by
	\begin{equation}
	x_{k+1} = F(x_k) = x_k + \int_{t_k}^{t_k+\Delta t} f(x(\tau))d\tau,
	\label{eq:sys}
	\end{equation}
	where $x \in \pazocal{M}$ is an $n$-dimensional state evolving on a smooth manifold according to the flow map $F: \pazocal{M} \rightarrow \pazocal{M}$, which can be related to an analogous continuous-time system $f(x)$ by the discretization shown above. 
	
	The Koopman operator is an infinite-dimensional linear operator capable of describing the evolution of any measure-preserving dynamical system, through its action on system observables \cite{original_koopman}. The operator describes the evolution of observables $g: \pazocal{M} \rightarrow \mathbb{R}$, which are elements of an infinite-dimensional Hilbert space. Although typically observables are taken from the space of Lebesgue square-integrable functions, other $L^p$ measure spaces are valid as well \cite{brunton_invariantsubspaces}. The action of the infinite-dimensional Koopman operator, $\pazocal{K}$, on an observable $g$ is given by
	\begin{equation}
	g(x_{k+1}) = g(F(x_k)) = \pazocal{K}g(x_k).
	\end{equation}
	
	Despite their infinite-dimensionality, Koopman operators can be approximated in finite dimensions, and used to describe nonlinear dynamical systems as a result of the development of methods such as those in \cite{brunton_dmd}, \cite{williams_edmd}. Given a dataset $X = [x_0,...,x_M]$ consisting of a single time-series of observations from a realization of a dynamical system, one must first choose a set of basis functions with which to span some subspace of the underlying function space, $\{z_1(x),...,z_N(x)\}, \ s.t. \ z_i: \pazocal{M} \rightarrow \mathbb{R}, \ \forall i \in \{1,...,N\}$, where we can define $\psi(x) = [z_1(x),...,z_N(x)]^T, \ s.t. \ \psi: \pazocal{M} \rightarrow \mathbb{R}^N$ as a vector-valued function encompassing the action of all basis functions on a given system state.
	
	We want to develop a mapping between current states and their evolution from the time-series of observations $X$. By defining a transformed dataset $\Psi_{X} = [\psi(x_0),...,\psi(x_{M-1})]^T$, and its evolution $\Psi_{X'} = [\psi(x_1),...,\psi(x_{M})]^T$, we can describe such a mapping as $\Psi_{X'} = \Psi_XK+r(X)$, where $K$ is a finite-dimensional approximation of the infinite operator $\pazocal{K}$, and $r(X)$ is a residual error due to the approximation. We can minimize the residual $r(X)$ over the squared-error loss functional
	\begin{equation}
	\underset{K}{\text{min}} \ \frac{1}{2}\sum_{k=1}^{M-1}||\psi(x_{k+1})-K\psi(x_k)||^2,
	\label{eq:koop_opt}
	\end{equation}
	with closed-form solution
	\begin{equation}
		K = G^\dagger A,
		\label{eq:koop_sol}
	\end{equation}
	where $\dagger$ denotes the Moore-Penrose pseudoinverse and
	\begin{equation}
	G = \frac{1}{M}\sum_{k=0}^{M-1} \psi(x_k)\psi(x_k)^T, \ A = \frac{1}{M}\sum_{k=0}^{M-1} \psi(x_k)\psi(x_{k+1})^T.
	\label{eq:koop_sol_mats}
	\end{equation}
	We obtain a matrix $K \in \mathbb{R}^{N\times N}$ that is an estimate of the system dynamics over the observed domain of the data \cite{abraham_mbckoopman}. We will use Koopman operators to describe individual behaviors expressed in data.

	\begin{figure*}[h]
		\subfigure[Segmentation of $\theta(t)$ and $\dot{\theta}(t)$ trajectories of an optimal control solution to the cart-pendulum inversion problem with a sample system shown below.]{%
			\includegraphics[width=1.005\columnwidth]{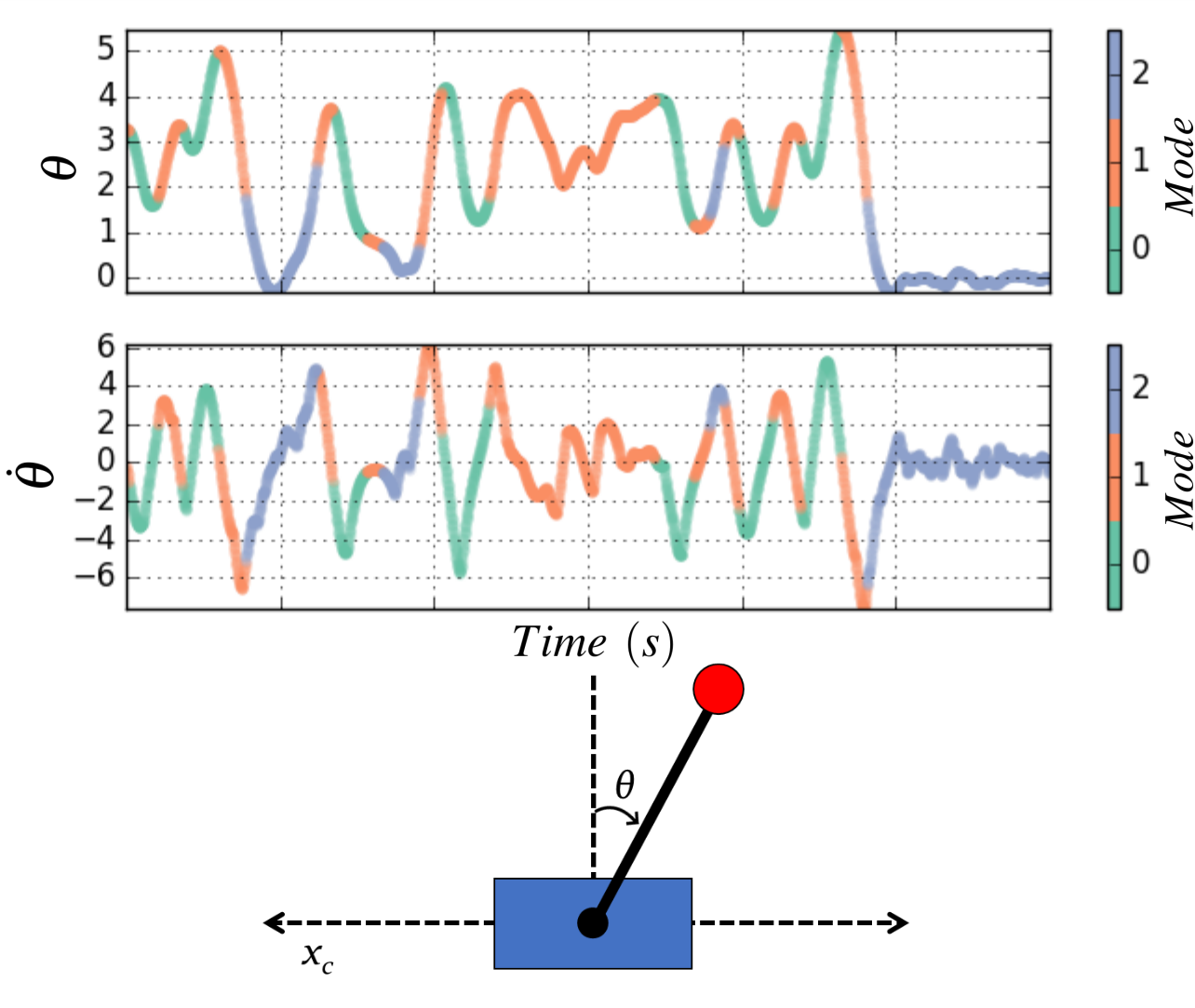} 
			\label{fig:sample_traj} 
		} 
		\quad
		\subfigure[Segmentation of the $(\theta,\dot{\theta})$ phase portrait of an optimal control solution to the cart-pendulum inversion problem.]{%
			\includegraphics[width=0.95\columnwidth]{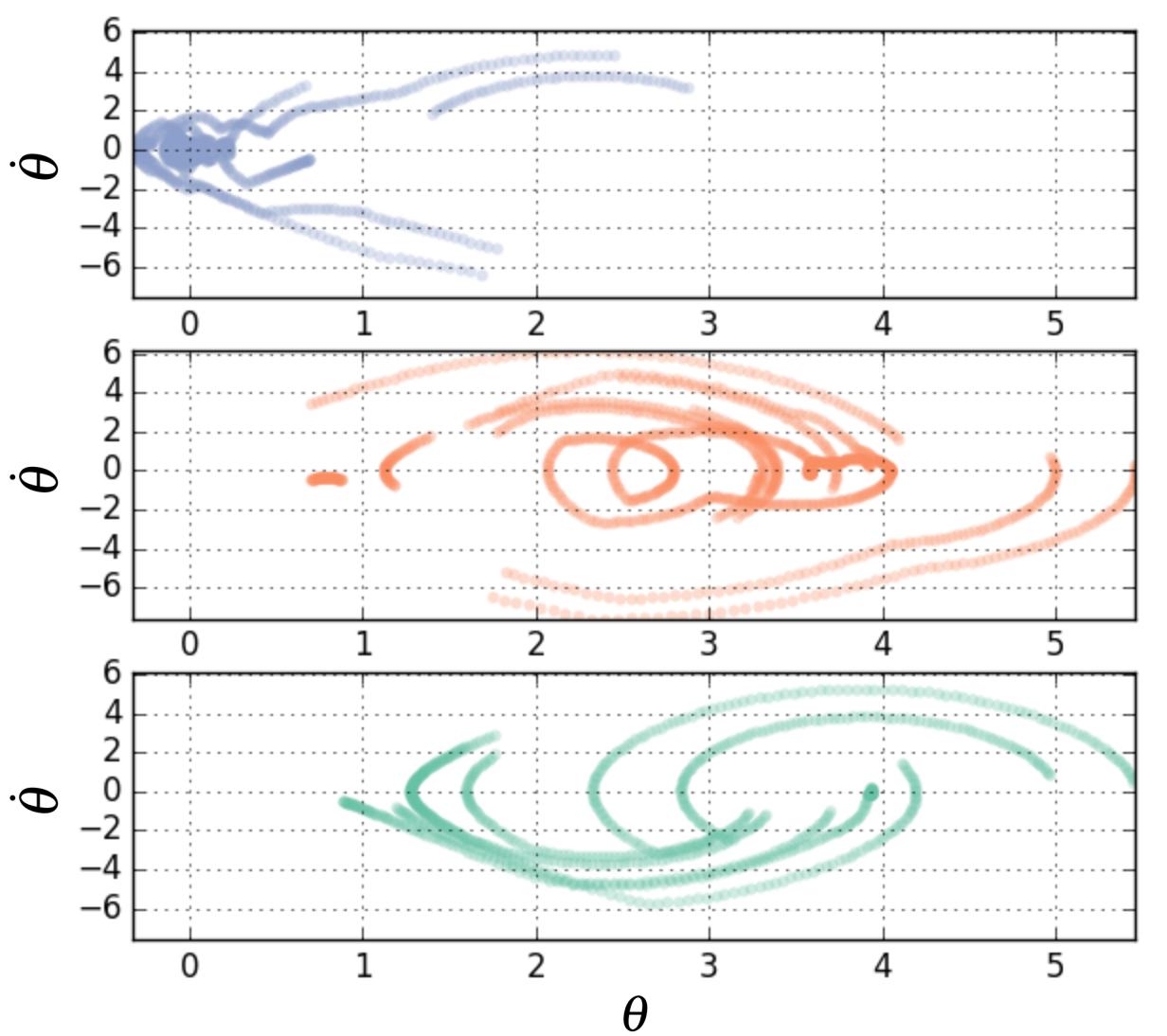} 
			\label{fig:sample_phase} 
		} 
		\caption{Example of dynamical system segmentation applied to an optimal model predictive control solution to the cart-pendulum inversion problem, specified by the goal state of $(\theta,x_c,\dot{\theta},\dot{x}_c)=(0,0,0,0)$. Despite the fact that the segmentation is not known \textit{a priori} it still corresponds to intuitive behaviors. The identified behaviors encode negative velocities, positive velocities, and low velocity motion in modes 0, 1 and 2 respectively.} 
	\end{figure*}

	\subsection{Dynamical System Segmentation}
	DSS characterizes all system behaviors over the state-space by synthesizing a non-redundant set of local estimates of the true system dynamics using a collection of Koopman operators. Given a dataset $X = [x_0,...,x_M]$ consisting of a single realization of a dynamical system of the same form as in Eq. \ref{eq:sys}, and a set of basis functions described by the vector-valued function $\psi(x) \ s.t \ \psi: \pazocal{M} \rightarrow \mathbb{R}^N$, we can apply the basis functions onto the dataset $X$ in order to generate a transformed dataset $\Psi_{X} = [\psi(x_0),...,\psi(x_{M})]^T$.	We then split the transformed dataset $\Psi_{X}$ into a set of $W+1$ overlapping rectangular windows, and calculate a Koopman operator for each, thereby generating a set of symbols $\mathbb{K} = \{K_0,...,K_W\}$. However, depending on the system under study, the size of the dataset, and choice of window size and overlap percentage, some of these symbols may be redundant.
	
	We are interested in creating a minimal alphabet of Koopman operators with which to span all system behaviors. Unsupervised learning methods such as clustering algorithms that specialize in the identification of classes within datasets are well-suited for this task. By considering each $\mathbb{R}^{N\times N}$ Koopman operator as a point in $\mathbb{R}^{N^2}$ space, we can divide the set $\mathbb{K}$ into subsets using a clustering algorithm. In particular, we use Hierarchical Density-Based Spatial Clustering of Applications with Noise (HDBSCAN), which is a nonparametric clustering algorithm that performs well in large databases subject to noise \cite{campello_hdbscan}. The algorithm groups the operators into $B+1$ classes $\{C_0,...,C_B\}$ using only the minimum of number of points required to make up a cluster as a parameter. We compose a set $\overline{\mathbb{K}}=\{\overline{K}_0,...,\overline{K}_B\}$ of class exemplars by taking a weighted-average of all $K_i \in C_j, \ \forall j \in \{0,...,B\}$, according to the class-membership probability $p(K_i|K_i \in C_j)$. The class-membership probability function is provided by the HDBSCAN software package \cite{McInnes2017}.
	
	Although we have created a minimal alphabet $\overline{\mathbb{K}}$ of system behaviors, it is of interest to project these behaviors onto the state-space manifold from this abstract operator space. We label all points in the transformed dataset $\Psi_X$ with a label $l\in \{0,...,B\}$ according to the class label of the Koopman operator each point was used to generate. Then, we train a support vector machine (SVM) classifier, $\Phi(\psi(x))$ to project the class labels onto the state-space manifold, thereby generating partitions of the state-space \cite{scikit-learn}.

\begin{figure}[!tp]
		\includegraphics[width=0.95\columnwidth,height=0.62\columnwidth]{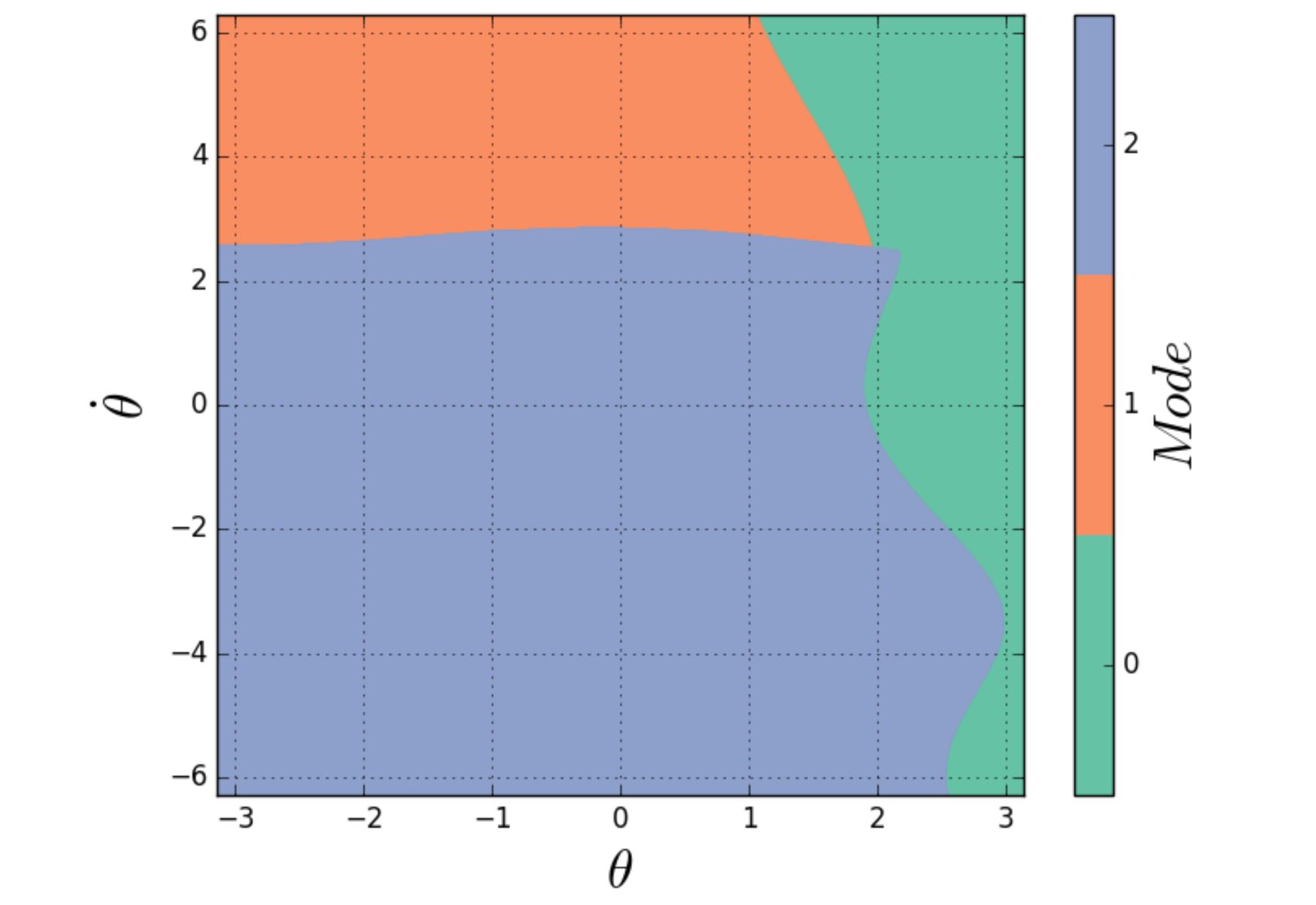}
		\caption{State-space partitions generated by an SVM trained on an optimal controller's cart-pendulum inversion. DSS identified 3 modes, and the SVM partitions are shown at the $(x_c,\dot{x}_c) = (1,-1)$ cross-section of the manifold.}
		\label{fig:phase_partition}	
	\end{figure}

	Figures \ref{fig:sample_traj} \&  \ref{fig:sample_phase} depict a cart-pendulum system used for an example application of DSS, where we segment an optimal control solution to the pendulum inversion task. Often the partitions generated by dynamical system segmentation may be intuitively related to the examined task. In Fig. \ref{fig:sample_traj}, one can see that modes 0 and 1 represent trajectories with negative and positive velocities respectively, while mode 2 represents lower velocity motion and stabilization. Since the state-space trajectories used to train the model encode task-specific information, the behavioral modes do as well. Once a dynamical system has been segmented, the SVM's partitions of the state-space are set, and new data points will be classified according to which partition they fall into. Figure \ref{fig:phase_partition} shows a cross-section of the partitioned state-space manifold of the optimal controller solution to the cart-pendulum inversion shown in Fig. \ref{fig:sample_traj}.

	\subsection{Graphical Representation}
\begin{figure}[!pt]
		\includegraphics[width=\columnwidth]{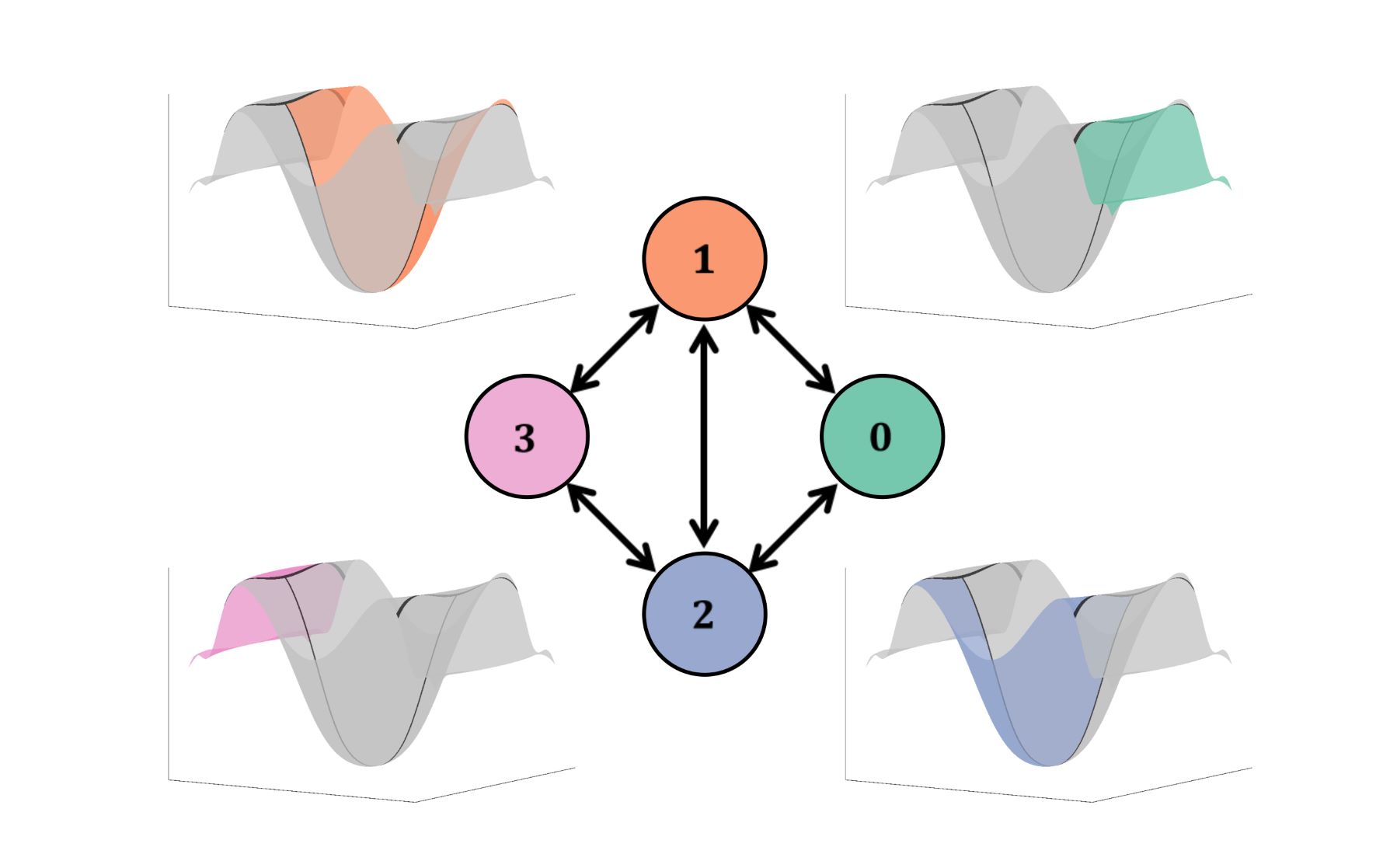}
		\caption{\textbf{Output of the Dynamical System Segmentation algorithm:} each node in the graph is a distinct dynamical system that governs its partition of the state-space manifold generated by the SVM $\Phi(\psi(x))$.}
		\label{fig:manifold}
	\end{figure}

	The product of DSS is best represented by a graph. We can define a graph $\mathbb{G} = (\overline{\mathbb{K}},\ \mathbb{E})$ where the node set $\overline{\mathbb{K}}$ contains the exemplar Koopman operators synthesized from the clustering procedure. The set of edges $\mathbb{E}$ is determined by directly observing the sequences of class labels in the dataset, and tracking all unique transitions. Figure \ref{fig:manifold} illustrates how DSS relates to the resulting graph. Each node in the graph represents a distinct dynamical system over its respective partition of the state-space manifold. By traversing the graph symbolically from one node to another, traversal of the state manifold is implied. The DSS algorithm is summarized in Algorithm 1. While initialization of a DSS model requires that the dataset $X$ be from a single realization of a system, additional realizations can be used to account for variability in system roll-outs.

	The graph itself encodes task-specific information embedded in the state trajectories of the training dataset. In particular, the graph's state distribution is an information-rich object that can be used for data analysis purposes. Given an optimal agent's graph $\mathbb{G}_{opt}$ constructed with DSS, we can use the trained SVM classifier $\Phi_{opt}(\psi(x))$ to identify behaviors from the optimal agent in data from other agents. By tracking the relative frequencies of behaviors $\overline{\mathbb{K}}$ from the optimal agent in another agent's trajectories, we can calculate a distribution $q(\overline{\mathbb{K}})$, and directly compare it to $\mathbb{G}_{opt}$'s optimal state distribution $p(\overline{\mathbb{K}})$ using the Kullback-Leibler divergence ($D_{KL}$) \cite{bishop_ml}
	
	\begin{algorithm}[!hpb]
		\caption{Dynamical System Segmentation (DSS)}
		\small
		\begin{algorithmic}[1]
			\renewcommand{\algorithmicrequire}{\textbf{Input:}}
			\renewcommand{\algorithmicensure}{\textbf{Procedure:}}
			\REQUIRE Dataset $X = [x_0,...,x_M]$ from a single realization of a dynamical system, basis functions $\{\psi(x)| \psi: \pazocal{M} \rightarrow \mathbb{R}^N\}$, window size $S_{w}$, overlap percentage $P_{ov}$, minimum number of points required to form a cluster $N_c$
			\ENSURE 
			\STATE Transform the $X$ dataset into $\Psi_{X} = [\psi(x_0),...,\psi(x_{M})]^T$
			\STATE Split $\Psi_X$ into $W+1$ windows of size $S_{w}$ overlapping by $P_{ov}$ 
			\STATE Calculate a Koopman operator $K_i$ for each window to construct the set $\mathbb{K} = \{K_0,...,K_W\}$
			\STATE Construct a feature array $\mathbb{K}_{flat}$ by flattening all $K_i \in \mathbb{R}^{N\times N}$ in $\mathbb{K}$ into points in $\mathbb{R}^{N^2}$ and appending them
			\STATE Cluster using HDBSCAN($\mathbb{K}_{flat}$, $N_c$), and label all $K_i$'s from one of $B+1$ discerned classes $\{C_0,...,C_B\}$
			\STATE Construct a set $\overline{\mathbb{K}}=\{\overline{K}_0,...,\overline{K}_B\}$ of class exemplars by taking a weighted-average of all $K_i \in C_j, \ \forall j \in \{0,...,B\}$, according to the membership probability $p(K_i|K_i \in C_j)$
			\STATE Label all points in $\Psi_X$ with the label $l \in \{0,...,B\}$ of the Koopman operator they were used to generate
			\STATE Train an SVM $\Phi(\psi(x))$ on the labeled points
			\STATE Construct a set of unique transitions $\mathbb{E}$ by tracking all sequential labels in the dataset
			\STATE Construct a graph $\mathbb{G} = (\overline{\mathbb{K}},\ \mathbb{E})$
			\renewcommand{\algorithmicensure}{\textbf{Return:}}
			\ENSURE Graphical model $\mathbb{G}$, and trained SVM $\Phi(\psi(x))$
			\label{alg:DSS}
		\end{algorithmic}
	\end{algorithm}

	\begin{equation}
		D_{KL}(p(\overline{\mathbb{K}})||q(\overline{\mathbb{K}})) = -\sum_{i=0}^{B}p(\overline{K}_i)log\big(\frac{q(\overline{K}_i)}{p(\overline{K}_i)}\big).
		\label{eq:KLdiv}
	\end{equation}
	
	The state distributions encode coarse-grained information about the task, and their comparison can be used for performance assessment. Since an optimal agent's distribution is the most representative state distribution of a task, $D_{KL}(p(\overline{\mathbb{K}})||q(\overline{\mathbb{K}}))$ represents the amount of task information embedded in an agent's motions, which we refer to as task embodiment. 

	\section{Experiments}
	The proposed assessment of task embodiment was applied to data collected from human subjects performing a cart-pendulum inversion task\footnote{The authors utilized de-identified data from a study approved by the Northwestern Institutional Review Board.}. Data was collected using the NACT-3D---an admittance-controlled haptic robot, similar to that described in \cite{stienen2011} and \cite{ellis2016}. We synthesize a dataset representative of an optimal user using an optimal controller. Data from the expert is segmented by applying the DSS algorithm proposed in Section II in order to generate a graphical model $\mathbb{G}_{opt}$, and a set of optimal behaviors to track. $\mathbb{G}_{opt}$'s state distribution is then used as a reference to compare against the human subjects, and assess their task embodiment.

	\subsection{Human Subjects Dataset}
\begin{figure}[!pb]
		\centering
		\includegraphics[width=0.95\columnwidth,height=.7\columnwidth]{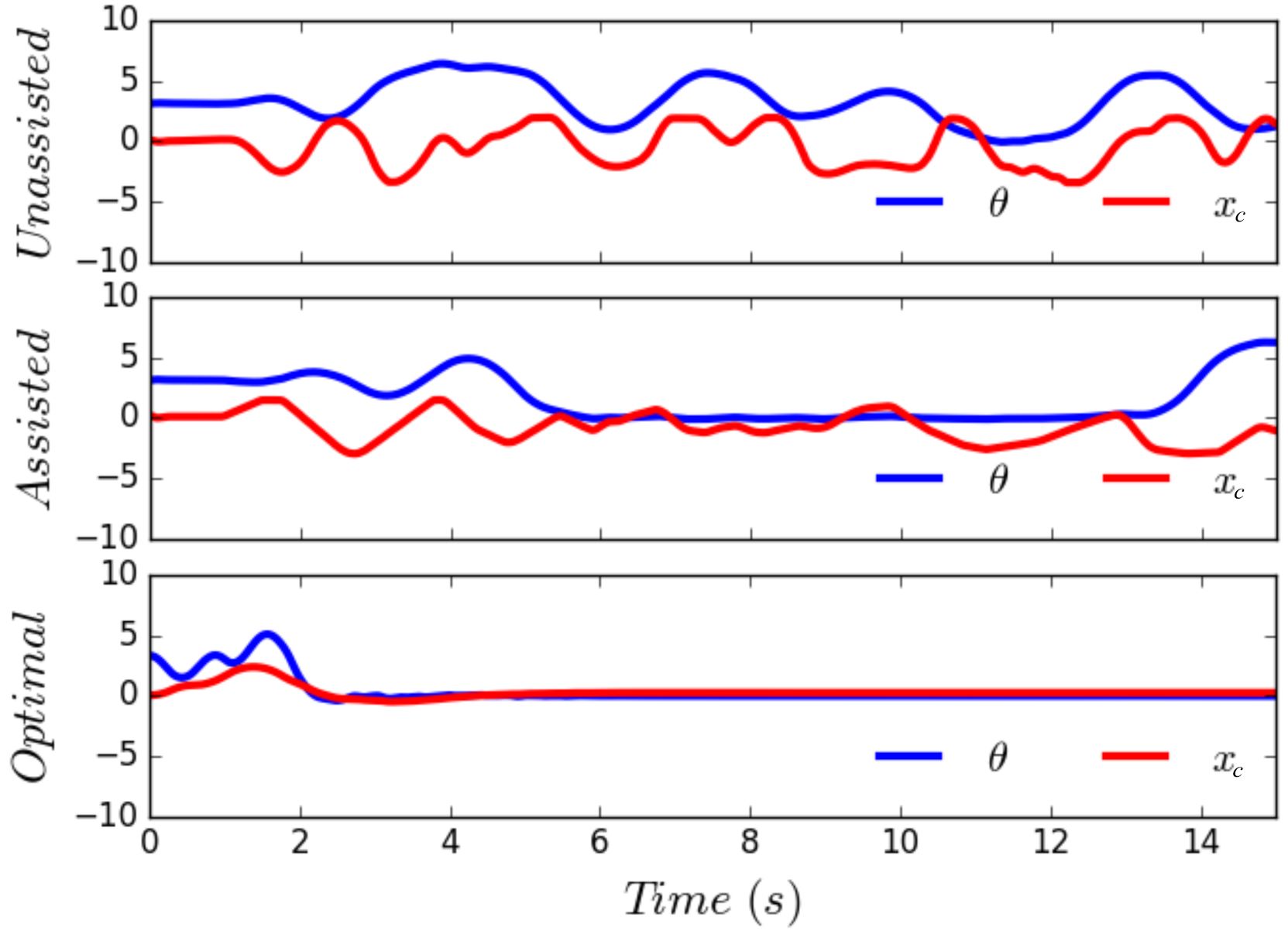}
		\caption{Sample trajectories from subject 16's trials depicting the effect of assistance in time-domain, as well as a trial from an optimal controller.}
		\label{fig:comparison}
	\end{figure}

	A filter-based assistance algorithm proposed in~\cite{therakis2015b} for pure noise inputs, and adapted for user input in~\cite{fitzsimons2016} and~\cite{kalinowska2018} was applied to a virtual cart-pendulum inversion task on the NACT-3D. The assistance physically filters the user's inputs---accelerations in this case---such that their actions are always in the direction of an optimal control policy calculated in real time. All subjects were instructed to attempt to invert a virtual cart-pendulum with the goal of spending as much time as possible in the unstable equilibrium during a thirty second trial, where the cart-pendulum states were sampled at $60Hz$. Subjects repeated this task for 30 trials in each of two sessions. Forty subjects completed this task with assistance in one session and without assistance in the other session. The order in which the subjects received assistance was counterbalanced to account for learning effects. An additional thirteen subjects were placed in a control group which completed both sessions without assistance. Figure \ref{fig:comparison} depicts the effect of assistance on the state trajectories of a representative subject, and includes an optimal trajectory for comparison. For the assisted trial, the subject reaches the goal state of $\theta=0$ and is able to balance the pendulum starting around $t=5s$. At this point, the assistance restricts the user's input motion $x_c(t)$ such that the inversion is maintained until $t=13s$. However, the same subject is unable to maintain the inverted configuration without assistance.

	\begin{figure*}[!htp]
		\subfigure[Time-domain segmentation of a selected optimal control solution of the pendulum inversion task. Mode 0 corresponds to energy pumping and swing-up, mode 1 corresponds to energy removal and slow-down, and mode 2 correponds to stabilization.]{%
			\includegraphics[width=1.0\columnwidth]{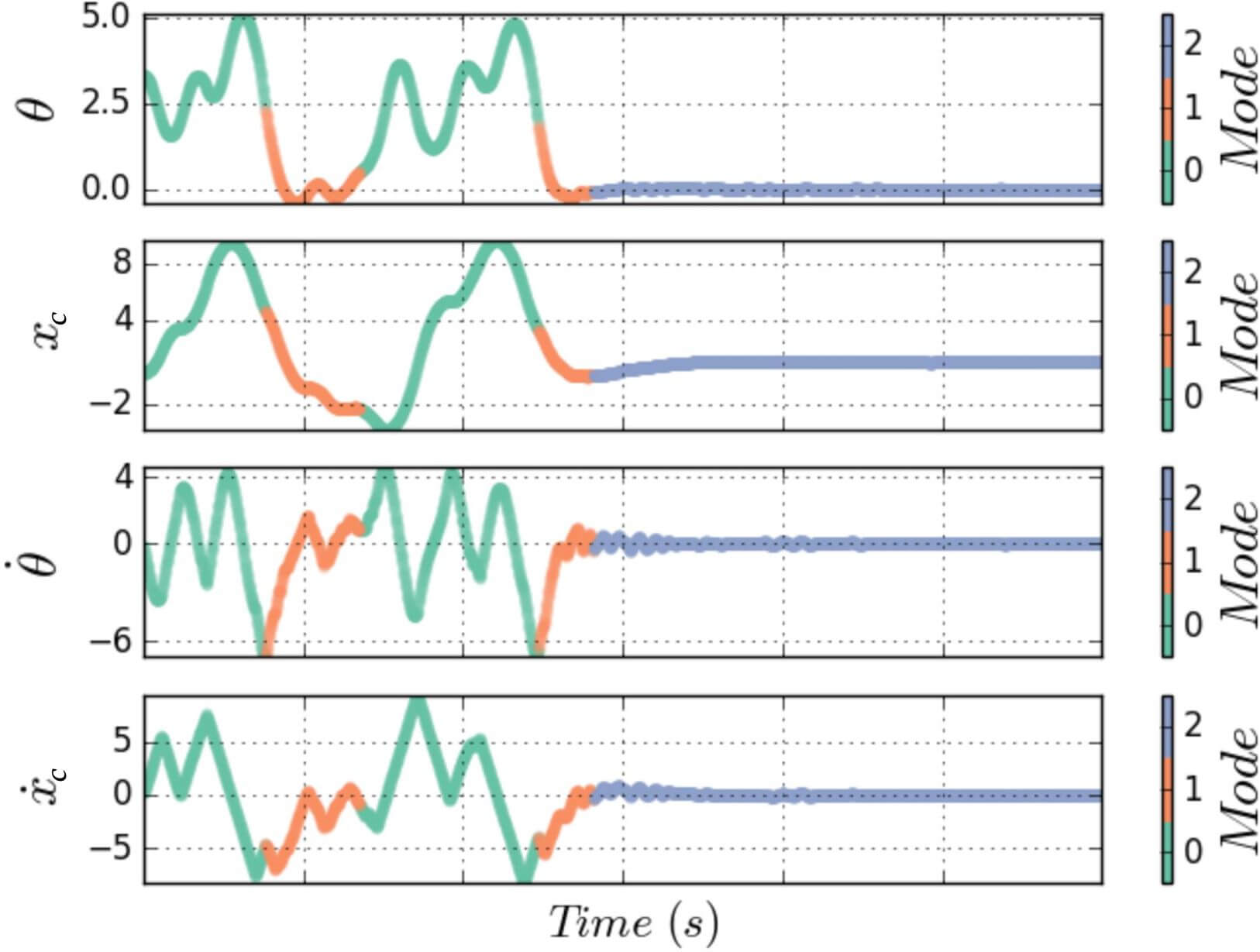} 
			\label{fig:opt_behaviors} 
		} 
		\quad 
		\subfigure[Graph $\mathbb{G}_{opt}$ resulting from the segmentation of a dataset of 30 optimal control solutions to the cart-pendulum inversion task. The set of segmented behaviors are shown projected onto the system's phase portrait over the domain $\{(\theta,\dot{\theta}):(-\pi,\pi) \times (-2\pi,2\pi)\}$.]{%
			\includegraphics[width=0.975\columnwidth]{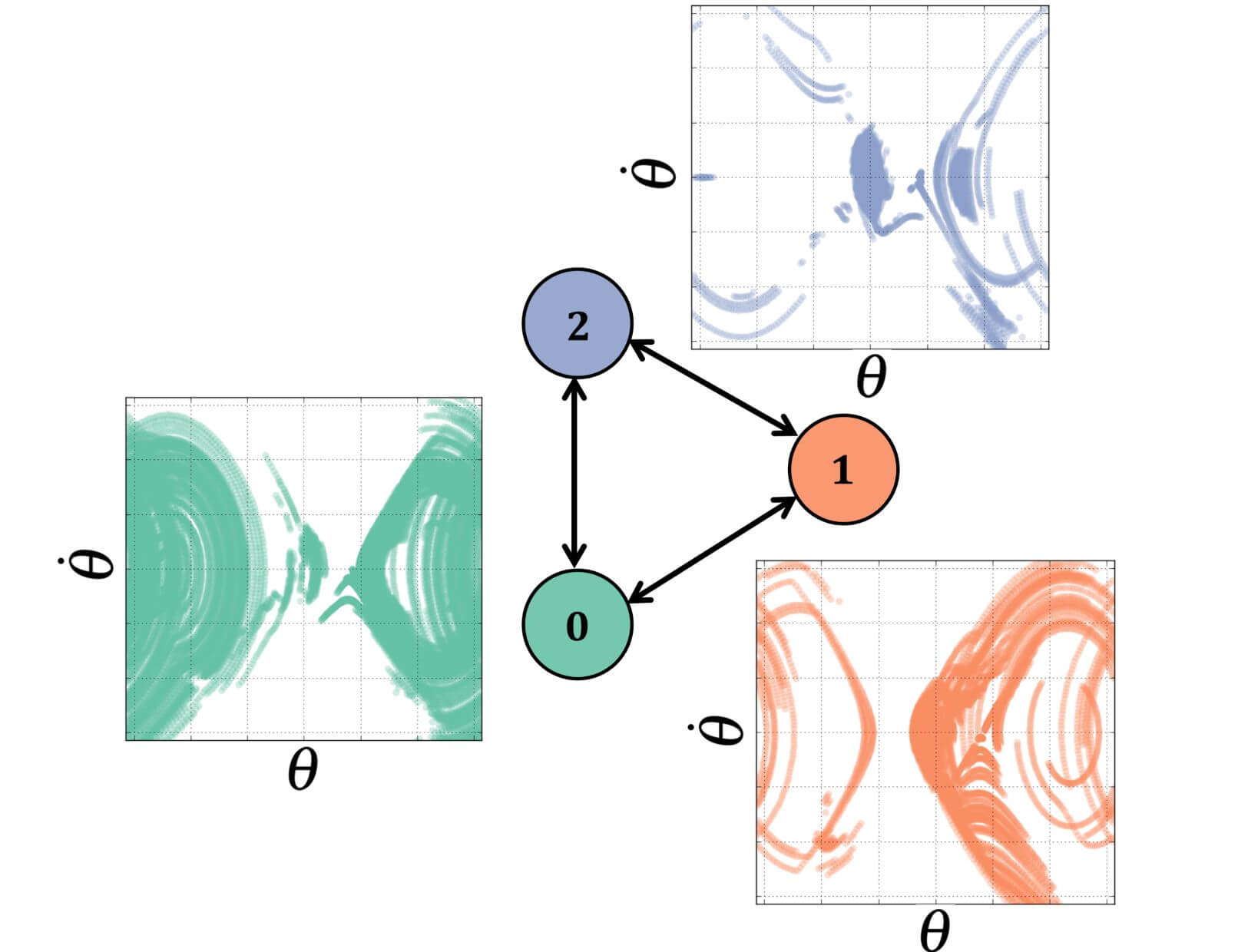} 
			\label{fig:dsac_manifold} 
		}
		\caption{Data-driven identification of exemplar behaviors through the use of dynamical system segmentation and the resulting graphical model from the segmentation of the pendulum inversion task.} 
	\end{figure*}

	\subsection{Training Dataset}
	To assess task embodiment using our dynamical system segmentation technique, we synthesized an optimal baseline to compare subjects against. We generated optimal control solutions to the pendulum inversion problem using Sequential Action Control \cite{SAC}, a receding-horizon model predictive optimal controller for nonlinear and nonsmooth systems, over a randomized set of initial conditions. The controller's objective was $(\theta,x_c,\dot{\theta},\dot{x}_c)=(0,1,0,0)$, with linear quadratic cost parameters of $Q=diag([200,80,0.01,0.2])$. Thirty optimal control trials of thirty seconds each were generated so as to mirror the amount of data collected from human subjects.

	\subsection{Optimal Graph}
	We apply the DSS algorithm to the synthesized trials to generate an optimal graphical model. The choice of basis functions has the greatest effect on the algorithm's performance because of how they reshape the state-space boundaries. The set of basis functions, $\psi(x)$, selected for this task were
	\begin{multline}
		\psi(x) = [\theta,\ x_c,\ \dot{\theta},\ \dot{x}_c,\ u,\ u\ cos(\theta),\ u\ cos(\dot{\theta}),
		\\
		|u_{sat}|cos^2\big(\frac{u \pi}{|u_{sat}|}\big),\ \dot{x}_c^2,\ 1],
	\label{eq:basis}
	\end{multline}
	where $|u_{sat}|$ is the optimal controller's saturation limit on the control effort. The basis functions were selected from the set of linear combinations of second order polynomial and sinusoidal functions. Since clustering occurs in $\mathbb{R}^{N^2}$ space, where $N$ is the number of basis functions, we chose a low-dimensional set ($N=10$) of representative basis functions in Eq. \ref{eq:basis} from the larger set of linear combinations of polynomial and sinusoidal functions. This dimensionality reduction can be achieved via multiple methods, such as principal component analysis \cite{bishop_ml}. 
	
	Figure \ref{fig:opt_behaviors} depicts the behaviors identified from the exemplar trial. The identified modes 0, 1 and 2 correspond to energy pumping and swing-up, energy removal and slow-down, and stabilization, respectively. These modes represent a set of behaviors that an expert user should exhibit in succeeding at the task.

	We synthesize the optimal graph $\mathbb{G}_{opt}$ using the identified behaviors, and then use the graph's state distribution, $p(\overline{\mathbb{K}})=[0.2437,\ 0.1275,\ 0.6288]$, as the reference baseline with which to assess the subjects' task embodiment. The graph $\mathbb{G}_{opt}$ and the segmented behaviors projected onto the $(\theta,\dot{\theta})$ phase portrait by the trained SVM $\Phi_{opt}(\psi(x))$ is shown in Fig. \ref{fig:dsac_manifold}.
	
	\begin{figure*}[!htp]
		\centering
		\includegraphics[width=.95\textwidth]{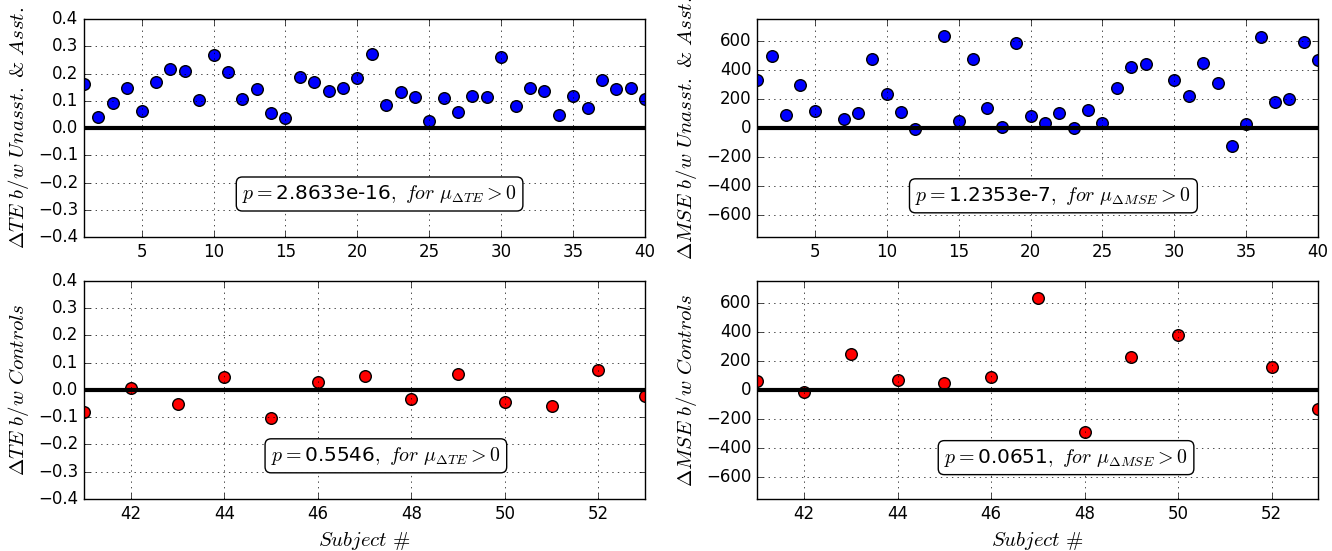}
		\caption{\textbf{Summary of experimental results:} subjects in the experimental group who received assistance (blue) were compared to their own unassisted trials. The control group subjects (red) were compared from their initial session to their final session. The pair of plots to the left show the difference in task embodiment between the sessions of the experimental and control groups. The plots to the right show the difference between the same groups using the integrated MSE instead. Both task embodiment and MSE are good predictors of assistance, validating task embodiment as a performance measure.}
		\label{fig:results}
	\end{figure*}

	The human data is analyzed by using the trained SVM $\Phi_{opt}(\psi(x))$ to detect the identified behaviors in each subject's trials with and without the presence of assistance. By tracking the relative frequencies of behaviors $\overline{\mathbb{K}}$ we can generate a distribution $q(\overline{\mathbb{K}})$ with which to compare to $\mathbb{G}_{opt}$'s state distribution $p(\overline{\mathbb{K}})$. We compare the distributions using task embodiment quantified by $D_{KL}(p(\overline{\mathbb{K}})||q(\overline{\mathbb{K}}))$, where a lower $D_{KL}$ indicates greater embodiment of the task. This same procedure is applied to the two sets of data from the control group subjects.

	\section{Results}
	We analyzed the human subjects dataset, and found that task embodiment is a reliable predictor of physical assistance. All subjects better embodied the task in their assisted trials, whereas there was no observed difference in the control group. In addition to comparing the groups using task embodiment, we also evaluated a standard metric for assessing task performance, the integrated MSE. Specifically, we calculated the integrated MSE with respect to a goal state of $(\theta,\dot{\theta})=(0,0)$. Integrated MSE is a reasonable performance metric for this task since success is defined as the ability to reach a single system configuration. However, we find that it predicts assistance at a lower significance level, and lower effect size than task embodiment.
	
	A paired two-sample t-test on the task embodiment of each subject with and without assistance showed that the subjects' sessions with assistance $(\mu=0.0756,\ \sigma=0.0436)$ significantly outperformed the sessions without assistance $(\mu=0.2084,\ \sigma=0.0560)$, with $p=\text{2.8633e-16}, \ t(39)=13.4876$, and an effect size of $d = 2.1326$. In contrast, there was no significant difference between the first session $(\mu=0.2039,\ \sigma=0.0406)$ and the second session $(\mu=0.1943,\ \sigma=0.0400)$ of the control group when a paired two-sample t-test was performed $p=0.5546, \ t(12)=-0.6051$. These results indicate that task embodiment reliably captures assistance and lack thereof.
	
	We also performed a paired two-sample t-test on the MSE of each subject with and without assistance, and found that the session with assistance $(\mu=124.66,\ \sigma=119.96)$ significantly outperformed the session without assistance $(\mu=428.88,\ \sigma=307.46)$, but with a lower significance and effect size than task embodiment, with $p=\text{1.2353e-7}, \ t(39)=6.4526$ and an effect size of $d = 1.0202$. Again, we applied the paired two-sample t-test to the control group and found that the first session $(\mu=352.83,\ \sigma=217.67)$ did not significantly outperform the second $(\mu=546.31,\ \sigma=446.10)$, had $p=0.0651, \ t(12)=2.0320$. These results indicate that MSE can also predict the presence of assistance, but not as reliably as task embodiment. The task embodiment measure has both a significance level several orders of magnitude greater than that of integrated MSE, and showed an effect size that was twice as large as integrated MSE. This demonstrates that task embodiment captures the large difference between the assisted and unassisted trials. These results are summarized in Fig. \ref{fig:results}, where we see that the change in task embodiment ($\Delta TE$) from assisted to unassisted trials is always positive. When we limit our alphabet to linear symbols using the same DSS hyperparameters as those in the outlined results, $\Delta TE$ was positive only in $5\%$ of subjects. When we tune the DSS parameters to generate an alphabet of the same size as in the primary results, $\Delta TE$ was positive only in $30\%$ of subjects. Therefore, for this system, linear symbols do not capture sufficient task information to reveal the presence of assistance.

\begin{figure*}[!htp]
		\subfigure[Resulting graph and state-space projections from the segmentation of subject 16's assisted trials of the pendulum inversion task.]{%
			\includegraphics[width=\columnwidth]{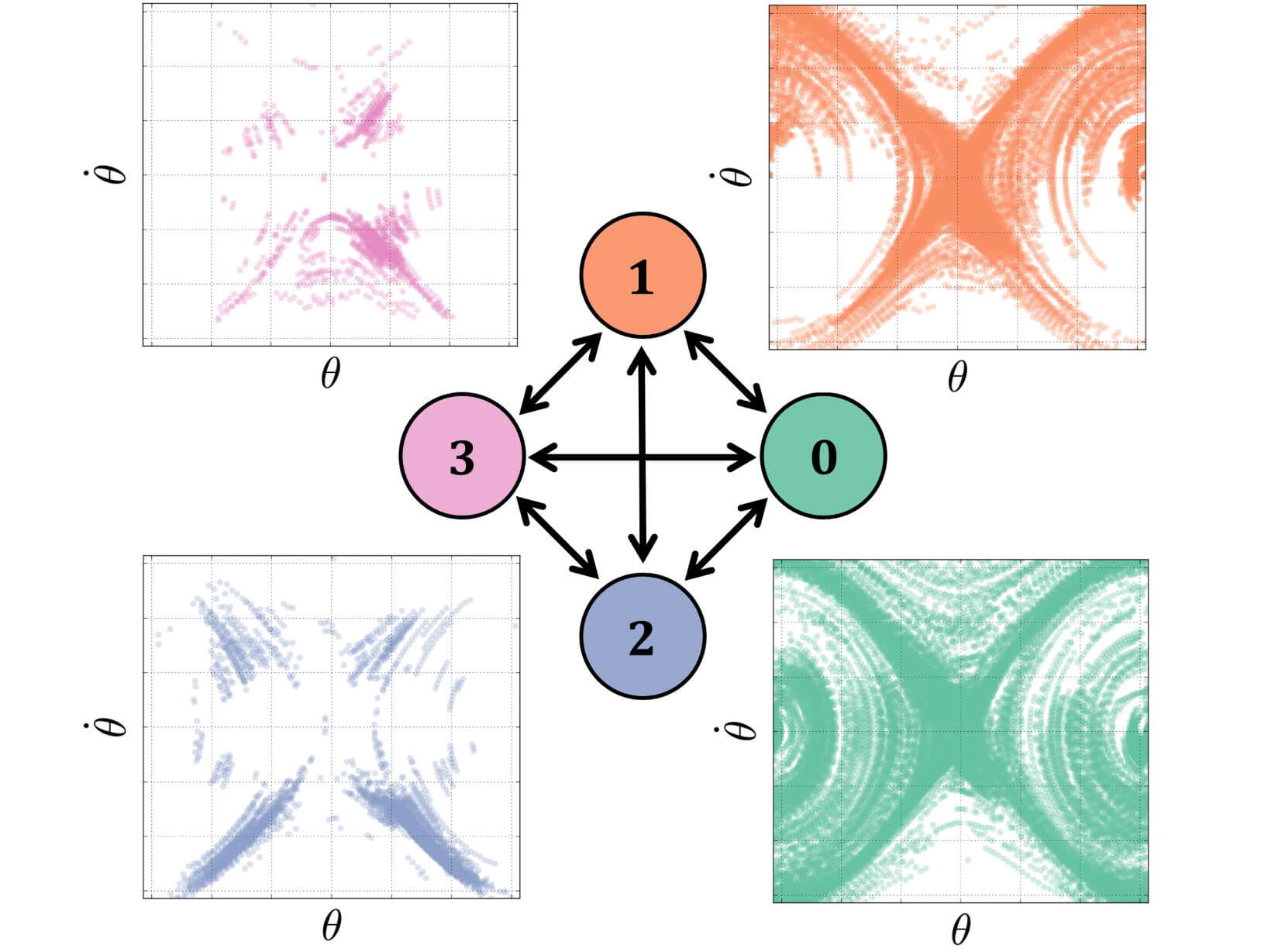} 
			\label{fig:s16mda_manifold} 
		} 
		\quad
		\subfigure[Resulting graph and state-space projections from the segmentation of subject 16's unassisted trials of the pendulum inversion task.]{%
			\includegraphics[width=0.99\columnwidth]{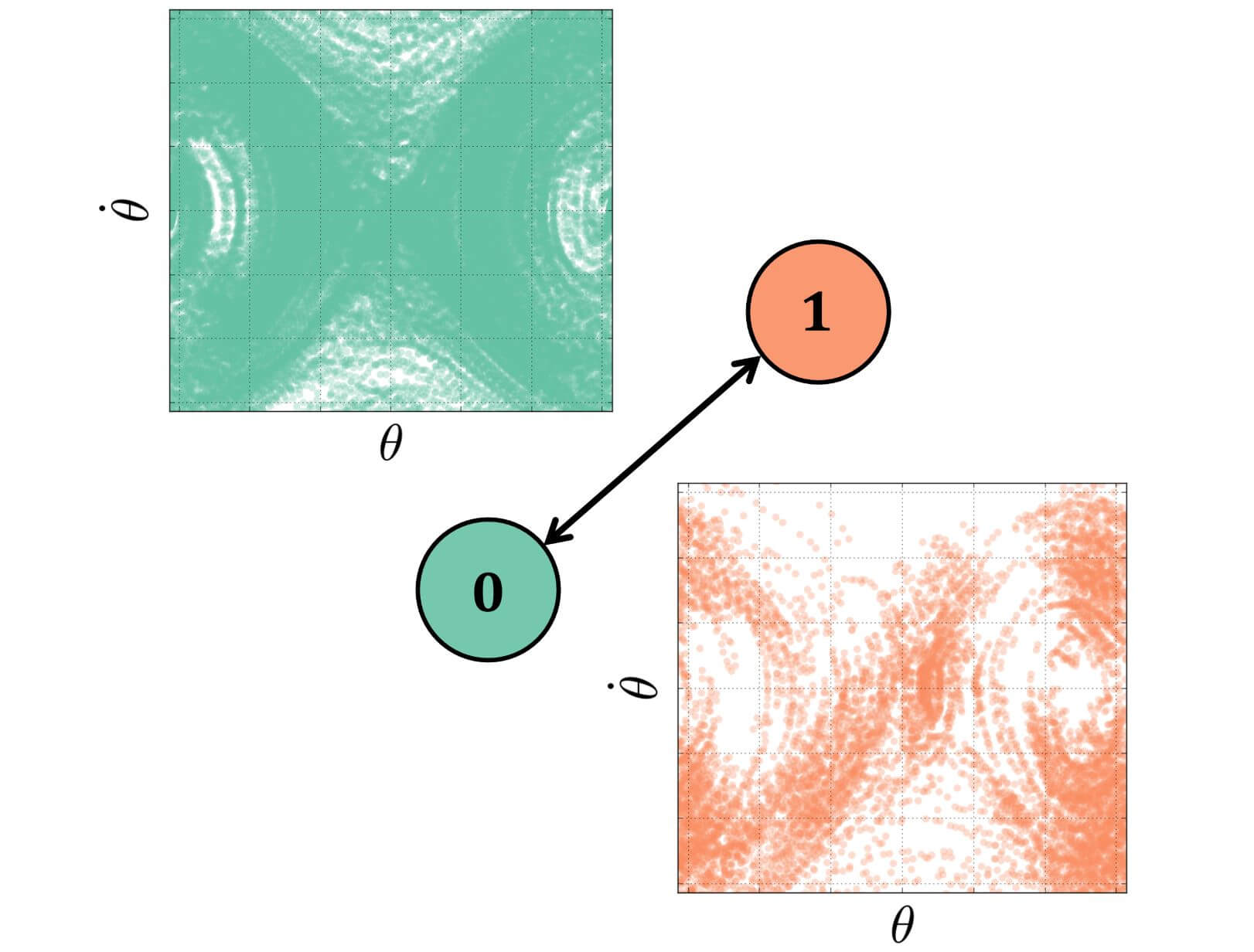} 
			\label{fig:s16nomda_manifold} 
		} 
		\caption{Constructed graphical models from the segmentation of a representative experimental subject with and without assistance. The graph's nodes project onto the system's $(\theta,\dot{\theta})$ phase portrait over the domain $\{(\theta,\dot{\theta}):(-\pi,\pi)\times(-2\pi,2\pi)\}$ of the state-space manifold according to the phase portraits shown alongside the nodes. We note that the behaviors of the unassisted subject's phase portraits are noise-driven and show no discernible structure.} 
	\end{figure*}

	The experimental methodology presented in this study analyzed subject data by means of comparison to an optimal baseline. While the methodology is informative, it cannot detail subject performance without comparison to the optimal agent. Given the same choice of basis functions and algorithm parameters, we can use DSS to generate graphs of each subject with and without assistance, and analyze the identified behaviors in each graph directly. This alternative methodology allows us to take human motion data and represent it graphically, which creates the opportunity for analyzing human motion using graph-theoretic principles. Figures \ref{fig:s16mda_manifold} \& \ref{fig:s16nomda_manifold} illustrate the graphical models constructed from the assisted and unassisted trials of a representative subject. We note that the extracted behaviors from the unassisted trials in Fig. \ref{fig:s16nomda_manifold} lack structure, and more closely resemble noise-driven behaviors. In contrast, by inspecting the graph from the subject's assisted trials in Fig. \ref{fig:s16mda_manifold}, we observe the emergence of finite structure in the identified behaviors.
	
	\section{Conclusions}
	In this study, we proposed an information-theoretic approach to human motion analysis. The DSS algorithm formulated in Section II produces graphical models that encode task-specific information. By tracking the degree of task embodiment, we are able to decode complex relationships in human motion. We applied DSS to a dataset of human subjects performing a virtual cart-pendulum inversion task with and without assistance. We determined that task embodiment is a good predictor of assistance, and validated the results by comparing task embodiment to integrated MSE. Moreover, task embodiment identified the presence of task assistance at a higher significance level and with a larger effect size than integrated MSE. Thus, the experimental results provide strong support for the use of information measures in human motion analysis.

	
\end{document}